\documentclass[runningheads]{llncs}
\usepackage[table]{xcolor}
\usepackage{colortbl}
\usepackage{amsmath}
\usepackage{ulem}
\usepackage{array}
\usepackage{multirow}
\usepackage{fancybox}
\usepackage{amssymb}
\usepackage{algorithm}
\usepackage{algorithmic}
\usepackage{graphicx}
\usepackage{booktabs}
\usepackage{xcolor}
\usepackage{colortbl}

\newcolumntype{L}[1]{>{\raggedright\let\newline\\\arraybackslash\hspace{0pt}}p{#1}}
\newcolumntype{P}[1]{>{\centering\let\newline\\\arraybackslash\hspace{0pt}}p{#1}}

\begin{document}
\title{Following the Teacher's Footsteps: Scheduled Checkpoint Distillation for Domain-Specific LLMs}
%
%
\author{Cheng Feng\inst{1}\orcidID{0000-0003-3861-0042} \and
        Chaoliang Zhong\inst{1}\orcidID{0000-0003-1697-6932} \and \\
        Jun Sun\inst{1}\orcidID{0000-0002-0967-4859} \and
        Yusuke Oishi\inst{2}\orcidID{0000-0003-4264-8932}}
\authorrunning{C. Feng et al.}
%
\institute{Fujitsu R\&D Center, Beijing, China\\
          \email{\{fengcheng, clzhong, sunjun\}@fujitsu.com}
          \and
          Fujitsu Research, FUJITSU LTD, Japan\\
          \email{oishi.y@fujitsu.com}}
\maketitle              
\begin{abstract}
Large language models (LLMs) are challenging to deploy for domain-specific tasks due to their massive scale. While distilling a fine-tuned LLM into a smaller student model is a promising alternative, the capacity gap between teacher and student often leads to suboptimal performance. This raises a key question: when and how can a student model match or even surpass its teacher on domain-specific tasks? In this work, we propose a novel theoretical insight: a student can outperform its teacher if its advantage on a Student-Favored Subdomain (SFS) outweighs its deficit on the Teacher-Favored Subdomain (TFS). Guided by this insight, we propose Scheduled Checkpoint Distillation (SCD), which reduces the TFS deficit by emulating the teacher’s convergence process during supervised fine-tuning (SFT) on the domain task, and a sample-wise Adaptive Weighting (AW) mechanism to preserve student strengths on SFS. Experiments across diverse domain tasks—including QA, NER, and text classification in multiple languages—show that our method consistently outperforms existing distillation approaches, allowing the student model to match or even exceed the performance of its fine-tuned teacher.

\end{abstract}
\section{Introduction}
Large language models (LLMs) have demonstrated remarkable few-shot learning capabilities~\cite{brown2020language,chowdhery2023palm,zhang2022opt}. However, their enormous model size poses substantial challenges for real-world deployment. For instance, performing inference on a 175-billion-parameter LLM requires at least 350GB of GPU memory even with specialized hardware~\cite{zheng2022alpa}. These extreme resource demands render such models prohibitively expensive for most domain-specific applications, which only require excellence on specific tasks, not universal capability.

\begin{figure}[t]
\includegraphics[width=\textwidth]{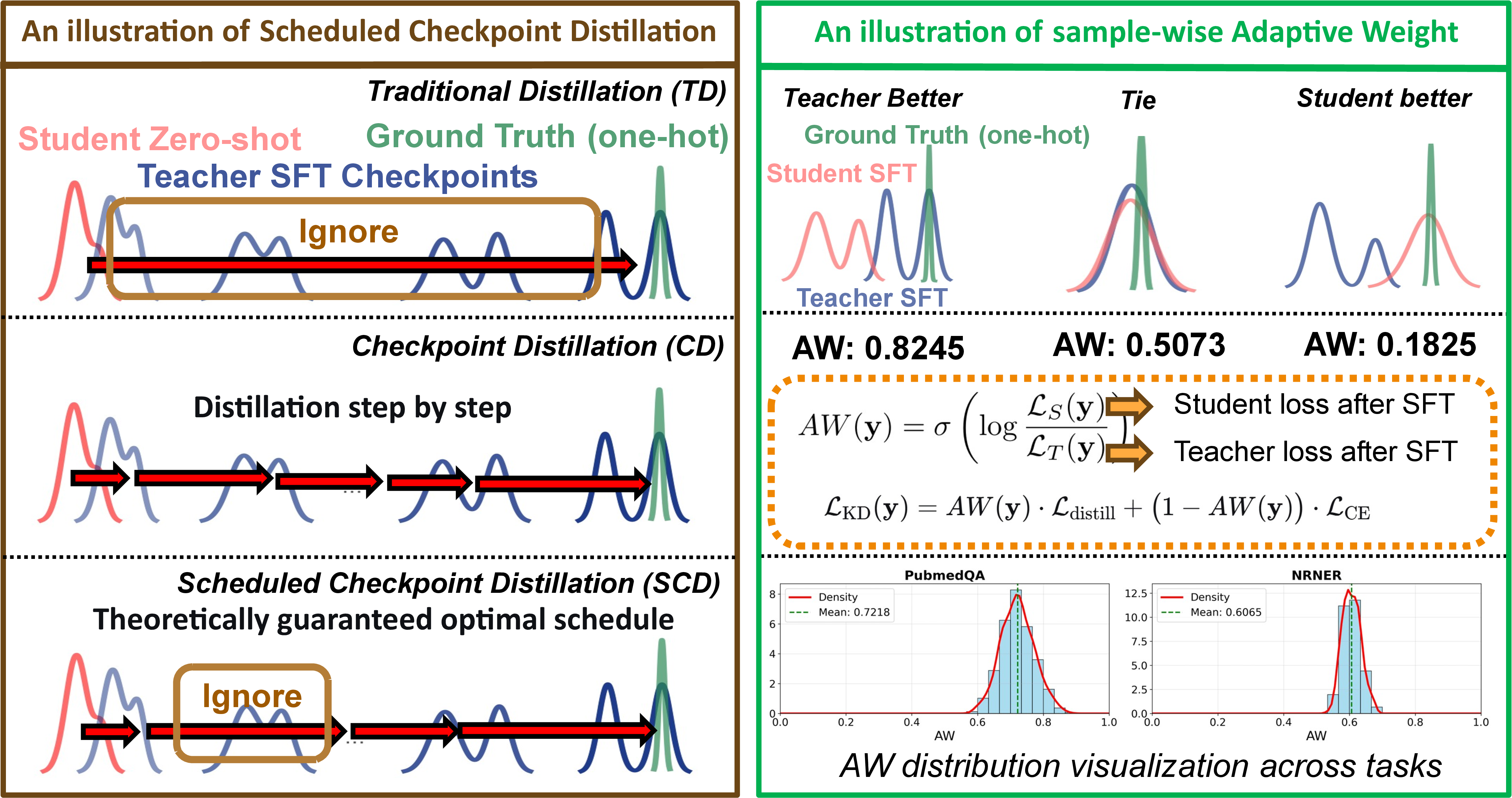}
\caption{On the Visualization of the proposed methods: Tracking Sample-wise Output Distribution Evolution. Scheduled Checkpoint Distillation (SCD, left) and sample-wise Adaptive Weighting (AW, right). SCD mimics the teacher's training trajectory to ease student learning. AW calibrates distillation weights per sample based on post-SFT output distributions, enhancing transfer where the teacher excels and suppressing it where the student is proficient.} \label{introduction_figure}
\end{figure}

A promising strategy for designing compact, high-performance domain-specific models involves domain adaptation of a large LLM via supervised fine-tuning (SFT), followed by compression through knowledge distillation (KD) ~\cite{hinton2015distilling} for deployment. KD facilitates the transfer of knowledge—particularly in the form of output distributions—from a thoroughly trained, high-capacity teacher model to a more lightweight student model. As a result, the student frequently attains superior performance compared to domain-specific SFT~\cite{liu2024ddk,kang2025distilling}. Nevertheless, the capacity gap in KD fundamentally constrains the student, hindering knowledge transfer and often causing failures to capture the teacher's complex output distribution—resulting in either oversimplified (mode averaging) or over-specialized (mode collapse) outcomes~\cite{gu2023minillm}. In this paper, we focus on the domain-specific SFT-then-distill pipeline and address its key challenge:

\textbf{\textit{When and how the student model can match, or even surpass, the teacher's performance on a domain-specific task?}}

We partition the target domain into student-favorable subdomain (SFS) where student loss is smaller and teacher-favorable subdomain (TFS). This division, while ubiquitous, manifests with heterogeneous distributions across different domain-specific tasks. This division stems from differences in their foundational expertise and subsequent optimization paths. Based on the distillation generalization theory~\cite{lopez2015unifying}, we show that \textbf{\textit{the student can match or surpass the teacher if and only if its advantage in the SFS outweighs its approximation deficit in the TFS}}. Full derivations are in Section 4.1.

Based on the above findings, we propose our methods from two aspects. First, we reduce the approximation deficit by improving upon the SOTA method, i.e., Checkpoint Distillation (CD)~\cite{panigrahi2024progressive}. Our key improvement is a theoretically grounded optimal checkpoint schedule, which contrasts with its original step-wise approach, as shown in Figure~\ref{introduction_figure} (left). Second, we seek to preserve the student's inherent strengths in SFS to ultimately surpass the teacher. Building on the sample-wise SFS and TFS distributions derived from the domain-specific SFT student and teacher models, we implement an instance-level Adaptive Weighting (AW) mechanism, as shown in Figure~\ref{introduction_figure} (right).  

To evaluate our approach, we conduct extensive experiments across diverse domain-specific tasks—including Question Answering (QA), Named Entity Recognition (NER), and Text Classification—in multiple languages. Results demonstrate that our method consistently surpasses SOTA distillation techniques. Notably, the distilled student model even outperforms the fine-tuned teacher in several tasks. These findings confirm our approach’s practicality and robustness, offering a promising path for efficient deployment in resource-aware scenarios. The main contributions are:
\begin{enumerate}
    \item We provide a novel theoretical analysis of when and how students can match or surpass their teachers in the \textsc{SFT}-then-distill pipeline.    
    \item Based on the theoretical results, we introduce a theoretically grounded selection strategy to reduce the approximation deficit caused by the teacher--student capacity gap, i.e., scheduled checkpoint distillation (SCD).    
    \item We propose a novel strategy, i.e., AW for retaining the student's advantages on \textsc{SFS} while absorbing the teacher's strengths on \textsc{TFS} through distillation.
\end{enumerate}

\section{Related Work}
\textbf{On bridging the teacher-student gap}: Previous work attributes this challenge to the teacher-student capability gap, mitigated via intermediate supervision using multi-step distillation~\cite{mirzadeh2020improved,harutyunyan2023supervision,jafari2021annealing} or trajectory guidance~\cite{shi2021follow}. Among these, \cite{harutyunyan2023supervision} shows the benefit of the teacher's intermediate training states. CD~\cite{panigrahi2024progressive} identifies this process as an implicit curriculum—accessible only via intermediate checkpoints—as a key gap-bridging mechanism and argues that a well-selected intermediate checkpoint, even if less accurate, yields a stronger student. However, what constitutes a "well-selected teacher" remains underexplored. In this paper, we deepen this discussion and reveal that optimal teacher selection requires a trade-off between the teacher's generalization and its capability gap. 

\textbf{On sample-wise adaptive weighting}: Current distillation methods typically use a fixed ratio (e.g., 0.5/0.5) to balance SFT and distillation losses, but such static weighting is suboptimal per sample due to gradient conflict, gradient dominance~\cite{openreview2025dtokd}, and varying sample-wise distillation difficulty~\cite{li2025sample}. To mitigate this, recent methods~\cite{openreview2025dtokd,li2025sample} dynamically adjust weights by tracking sample-level statistics during training, yet introduce significant overhead and instability as the sample states changes over training. We address this via a lightweight static weighting strategy, which assigns difficulty-aware weights using only one forward pass through both teacher and student SFT models—without dynamic adjustment or training-time cost.

\section{Preliminaries}

\subsection*{Problem Setting for Domain-Specific LLM Distillation}

Let a language model be defined as a probability distribution $p$ over token sequences $\mathbf{y} = (y_1, y_2, \dots, y_S) \in \mathcal{Y}^S$, where $\mathcal{Y}$ is the vocabulary and $S$ is the sequence length. The distribution is obtained by applying the softmax function to the model's logits:
\[
p(y_s \mid \mathbf{y}_{<s}) = \frac{\exp\bigl(\mathrm{logit}_p(y_s \mid \mathbf{y}_{<s})\bigr)}{\sum_{y' \in \mathcal{Y}} \exp\bigl(\mathrm{logit}_p(y' \mid \mathbf{y}_{<s})\bigr)},
\]
and satisfies the autoregressive property:
\[
p(\mathbf{y}) = \prod_{s=1}^{S} p(y_s \mid \mathbf{y}_{<s}), \qquad \mathbf{y}_{<s} := (y_1, \dots, y_{s-1}),
\]
with $p(y_1 \mid \mathbf{y}_{<1}) = p(y_1)$ for $s = 1$.

In domain-specific knowledge distillation, we first obtain a teacher model $p_T$ by performing supervised fine-tuning (SFT) of a foundation model on the target domain dataset $\mathcal{D}_{\mathrm{domain}}$. Knowledge is then transferred from $p_T$ to a smaller or more efficient student model $q\theta$. The overall objective is to find parameters $\theta$ that minimize a compound loss $\mathcal{L}(\theta)$, which jointly optimizes a distribution divergence from the teacher, i.e., $\mathcal{L}_{\text{distill}}$ and alignment with ground-truth labels, i.e., the cross-entropy loss $\mathcal{L}_{\text{CE}}$:
\[
\mathcal{L}(\theta) = \alpha \cdot \mathcal{L}_{\text{distill}}(p_T, q_\theta) + (1-\alpha) \cdot \mathcal{L}_{\text{CE}}(y_{\text{true}}, q_\theta),
\]
where $\alpha \in [0,1]$ is a balancing hyperparameter, commonly set to $0.5$ in practice.

\subsection*{Existing knowledge distillation approaches}
\textbf{Traditional Distillation (TD)} aims to transfer knowledge from a single well-trained teacher model to a smaller or more efficient student. Standard Knowledge Distillation (KD)~\cite{hinton2015distilling} employs the forward Kullback--Leibler (KL) divergence as the primary alignment objective:
\[
L_{\mathrm{KL}}(p,q_\theta)=\frac{1}{S}\sum_{s=1}^S \sum_{y_s\in\mathcal{Y}} 
p(y_s\mid\mathbf{y}_{<s})\log\frac{p(y_s\mid\mathbf{y}_{<s})}{q_\theta(y_s\mid\mathbf{y}_{<s})}.
\]
However, it is prone to the mode-averaging problem: the student may spread its probability mass too thinly across all modes of the teacher distribution. To tackle this, Wen et al.~\cite{wen2023f} introduced Reverse KL (RKL) divergence. Although this alleviates mode-averaging, it may cause mode collapse, concentrating the student solely on the teacher's dominant modes. Time‑Adaptive Intermediate Distillation (TAID)~\cite{shing2025taid} addresses this instability via curriculum-based logit-level interpolation. Instead of directly matching the teacher, the student is trained to follow a dynamically evolving intermediate target:
\[
p_t(y_s\mid\mathbf{y}_{<s}):=\mathrm{softmax}\!\bigl((1-t)\,\mathrm{logit}_{q'_\theta}+t\,\mathrm{logit}_{p}\bigr),
\]
where $t$ gradually increases from 0 to 1 during training, and $q'\theta$ denotes detached student logits, restricting gradients to flow only through $q\theta$ in the objective.
\[
L^{(t)}_{\mathrm{TAID}}(p,q_\theta):=L_{\mathrm{KL}}(p_t,q_\theta).
\]
\textbf{Checkpoint Distillation (CD)}~\cite{panigrahi2024progressive}

Let \(\mathcal{C}_T = \{ p^{(i)} \}_{i=1}^N\) denote the intermediate checkpoints saved during teacher SFT, where \(p^{(N)}\) is the final teacher. In checkpoint distillation, the student is trained progressively on a scheduled sequence of these checkpoints.

Formally, at training step \(k\), the student minimizes a distillation loss \(L_{\text{distill}}\) with respect to a scheduled checkpoint \(p^{(\tau_k)}\):

\[
\theta_{k+1} = \theta_k - \eta \, \nabla_{\theta} \, L_{\text{distill}}\!\bigl(p^{(\tau_k)}, q_{\theta_k}\bigr),
\]

where \(\tau: \mathbb{N} \to \{1,\dots,N\}\) defines the checkpoint schedule.

The common schedule is \((N,T)\)-progressive distillation:
\(\mathcal{C}_T\) contains \(N-1\) equally spaced intermediate checkpoints plus the final teacher. The student is trained sequentially with each checkpoint \(p^{(i)} \in \mathcal{C}_T\) for a fixed duration \(T\) steps, totaling \(NT\) steps before continuing solely with the final teacher \(p^{(N)}\). 

\section{Theoretical Analysis}
\subsection{Generalized Single Distillation}
According to David's theory~\cite{lopez2015unifying}, we get the following theorems:
\begin{theorem}[Student Generalization Bound]
Let $f_s \in \mathcal{F}_s$ be the student model and $f \in \mathcal{F}$ be the true target function. When learning from $n$ i.i.d. data samples, the student's excess risk is bounded by:
\[
\mathcal{R}(f_s) - \mathcal{R}(f) \leq O\left( \frac{|\mathcal{F}_s|_C}{\sqrt{n}} \right) + \epsilon_s,
\]
where $|\mathcal{F}_s|_C$ measures the complexity of the student's hypothesis class, $O(\cdot)$ is the statistical estimation error decaying at rate $1/\sqrt{n}$, and $\epsilon_s := \inf_{g \in \mathcal{F}_s} \mathcal{R}(g) - \mathcal{R}(f)$ is the approximation error of $\mathcal{F}_s$.
\end{theorem}

\begin{theorem}[Teacher Generalization Bound]
Let $f_t \in \mathcal{F}_t$ be the teacher model. Due to a better representation, the teacher achieves faster convergence with excess risk bounded by:
\[
\mathcal{R}(f_t) - \mathcal{R}(f) \leq O\left( \frac{|\mathcal{F}_t|_C}{n} \right) + \epsilon_t,
\]
where the estimation error decays at rate $1/n$, and $\epsilon_t := \inf_{g \in \mathcal{F}_t} \mathcal{R}(g) - \mathcal{R}(f)$ is the approximation error of $\mathcal{F}_t$.
\end{theorem}

\begin{theorem}[Student Distillation Generalization Bound]
 The student's risk relative to the teacher converges at an accelerated rate:
\[
\mathcal{R}(f_s) - \mathcal{R}(f_t) \leq O\left( \frac{|\mathcal{F}_s|_C}{n^{\alpha}} \right) + \epsilon_l, \quad \alpha > 0.
\]
Here $\alpha$ controls the distillation speed, and $\epsilon_l := \inf_{g \in \mathcal{F}_s} \left( \mathcal{R}(g) - \mathcal{R}(f_t) \right)$ is the distillation approximation error, representing the inherent limitation of $\mathcal{F}_s$ in approximating $f_t$.
\end{theorem}

\subsection{When and How Can Students Surpass Their Teachers}
To enable the student model to surpass its teacher, we aim for $\mathcal{R}(f_s) - \mathcal{R}(f_t) \le 0$. However, Theorem 3 cannot achieve such goal as $O\left( \frac{|\mathcal{F}_s|_C}{n^{\alpha}} \right) \ge 0$ and $\epsilon_l \ge 0$. Therefore, we partition the input domain $\mathcal{D}$ into two subdomains:

\begin{itemize}
    \item \textbf{Student-Favored (SFS)}: $\mathcal{D}_{\text{SFS}} := \{ x \in \mathcal{D} \mid \mathcal{R}_{x}(f_s) - \mathcal{R}_{x}(f_t) \le 0 \}$.
    \item \textbf{Teacher-Favored (TFS)}: $\mathcal{D}_{\text{TFS}} := \{ x \in \mathcal{D} \mid \mathcal{R}_{x}(f_s) - \mathcal{R}_{x}(f_t) > 0 \}$.
\end{itemize}

Here, $\mathcal{R}_{x}(f)$ denotes the point-wise or local risk of hypothesis $f$ at $x$. The global risk difference decomposes as:
\[
\mathcal{R}(f_s) - \mathcal{R}(f_t) = \underbrace{\left( \mathcal{R}_{\text{SFS}}(f_s) - \mathcal{R}_{\text{SFS}}(f_t) \right)}_{\le 0} + \underbrace{\left( \mathcal{R}_{\text{TFS}}(f_s) - \mathcal{R}_{\text{TFS}}(f_t) \right)}_{> 0}.
\]

\begin{theorem}[Sufficient Condition for Student Surpassing Teacher]
The sufficient condition for the student model $f_s$ to achieve lower total risk than the teacher $f_t$ is:
\[
\mathcal{R}_{\text{SFS}}(f_t) - \mathcal{R}_{\text{SFS}}(f_s) \ge O\left( \frac{|\mathcal{F}_s|_C}{n_{TFS}^{\alpha}} \right) + \epsilon_l^{\text{(TFS)}},
\]
where $\epsilon_l^{\text{(TFS)}} := \inf_{g \in \mathcal{F}_s} \left( \mathcal{R}_{\text{TFS}}(g) - \mathcal{R}_{\text{TFS}}(f_t) \right)$ is the local distillation approximation error on TFS and $n_{TFS}$ is the number of samples from TFS.
\end{theorem}
Our proposed method is also grounded in this theoretical insight: 
\begin{itemize}
    \item To \textbf{reduce the approximation deficit} ($O\left( \frac{|\mathcal{F}_s|_C}{n_{TFS}^{\alpha}} \right) + \epsilon_l^{\text{(TFS)}}$), we propose a \textit{scheduled checkpoint distillation} approach, which systematically leverages intermediate teacher checkpoints for alignment. Details shown in Section 5.
    
    \item To \textbf{enlarge the student's advantage} ($\mathcal{R}_{\text{SFS}}(f_t) - \mathcal{R}_{\text{SFS}}(f_s)$), we estimate the distributions of the SFS and TFS using domain-specific fine-tuned student and teacher models, and assign sample-wise weights accordingly. This weighting scheme enhances learning in regions where the student has potential for superiority. The methodology is elaborated in Section 6.
\end{itemize}

\section{Scheduled Checkpoint Distillation}
Our framework builds upon the standard CD~\cite{panigrahi2024progressive} setup, wherein a sequence of intermediate teacher checkpoints $\mathcal{C}_T = \{ p^{(i)} \}_{i=1}^N$ is saved during teacher fine-tuning, and the student is progressively trained on a subset of these checkpoints. We first ground checkpoint schedule design in theory. Based on this, we propose an algorithm to dynamically select the most informative checkpoints, targeting an \textit{optimal teacher checkpoint schedule}.


\subsection{Principles for Schedule}
\begin{theorem}[Individual Step Checkpoint Distillation Error Bound]
Consider the student model at iteration $i$, denoted as $f_s^{(i)}$, aligns with a selected teacher checkpoint $p^{(j)}$ from the checkpoint set $\mathcal{C}_T = \{ p^{(1)}, \dots, p^{(N)} \}$. The risk of the student after this alignment step satisfies:

\[
\mathcal{R}(f_s^{(i)}) \leq \mathcal{R}(p^{(j)}) + O\left( \frac{|\mathcal{F}_s|_C}{n^{\alpha}} \right) + \epsilon_{l}^{(i,j)},
\]

where: $\epsilon_{l}^{(i,j)} := \inf_{g \in \mathcal{F}_s} \left( \mathcal{R}(g) - \mathcal{R}(p^{(j)}) \right)$ . Let $f_s^{(i-1)} \in \mathcal{F}_s$ be the student model obtained at the previous step, it follows that:
\begin{align*}
\epsilon_{l}^{(i,j)} 
&:= \inf_{g \in \mathcal{F}_s} \left( \mathcal{R}(g) - \mathcal{R}(p^{(j)}) \right) \\
&= \inf_{g \in \mathcal{F}_s} \left\{ \left[ \mathcal{R}(g) - \mathcal{R}\bigl(f_s^{(i-1)}\bigr) \right] + \left[ \mathcal{R}\bigl(f_s^{(i-1)}\bigr) - \mathcal{R}(p^{(j)}) \right] \right\} \\
&\le \inf_{g \in \mathcal{F}_s} \left[ \mathcal{R}(g) - \mathcal{R}\bigl(f_s^{(i-1)}\bigr) \right] + \bigl| \mathcal{R}\bigl(f_s^{(i-1)}\bigr) - \mathcal{R}(p^{(j)}) \bigr| \\
&\le 0 + \bigl| \mathcal{R}\bigl(f_s^{(i-1)}\bigr) - \mathcal{R}(p^{(j)}) \bigr| \\
&\le \bigl| \mathcal{R}\bigl(f_s^{(i-1)}\bigr) - \mathcal{R}(p^{(j)}) \bigr|.
\end{align*}
\end{theorem}
Then we come to the final results that:
\[
\underbrace{\mathcal{R}(f_s^{(i)})}_{\text{Target}} 
\leq \underbrace{\mathcal{R}(p^{(j)})}_{\text{j-th teacher error}} 
+ \underbrace{O\left( \frac{|\mathcal{F}_s|_C}{n^{\alpha}} \right)}_{\text{Constant}} 
+ \underbrace{\bigl| \mathcal{R}\bigl(f_s^{(i-1)}\bigr) - \mathcal{R}(p^{(j)}) \bigr|}_{\text{Upper bound of } \epsilon_{l}^{(i,j)}}.
\]
The optimization problem is then transformed into find the $j^*$-th teacher for the $i$-th student where:
\[
j^* = \arg\min_{j \in N} \left[ \mathcal{R}(p^{(j)}) + \bigl| \mathcal{R}(f_s^{(i-1)}) - \mathcal{R}(p^{(j)}) \bigr| \right].
\]
Then the principles for schedule include the following key components:
\begin{itemize}
    \item \textbf{Select high-performance teachers}: Minimize $\mathcal{R}(p^{(j)})$ by choosing teacher with superior generalization ability.
    \item \textbf{Select proximal teachers}: Minimize $\bigl| \mathcal{R}(f_s^{(i-1)}) - \mathcal{R}(p^{(j)}) \bigr|$ by choosing teacher whose performance is close to the current untrained student's state.
\end{itemize}

\subsection{Schedule Metric}
Let $\mathcal{D}$ be the domain-specific input distributions. The teacher checkpoint schedule is designed based on two metrics:

\textbf{Metric 1. High-performance teacher:} 
Choose teacher checkpoints $p^{(j)}$ with minimal $\mathcal{R}(p^{(j)})$, where $\mathcal{R}(p^{(j)})$ could be evaluated over the domain $\mathcal{D}$.

\textbf{Metric 2. Proximal teacher:} 
Choose teacher checkpoints $p^{(j)}$ with minimal $\bigl| \mathcal{R}(f_s^{(i-1)}) - \mathcal{R}(p^{(j)}) \bigr|$, which can be approximated by the KL-divergence between student and teacher output distributions over $\mathcal{D}$:
\[
\bigl| \mathcal{R}(f_s^{(i-1)}) - \mathcal{R}(p^{(j)}) \bigr| 
\approx \mathop{\mathbb{E}}_{x \sim \mathcal{D}} \Big[ \text{KL}\big( f_s^{(i-1)}(x) \,\|\, p^{(j)}(x) \big) \Big].
\]
Since the two metrics are measured differently—Metric 1 is computed from sample loss, while Metric 2 is based on model outputs divergence—we harmonize them by expressing both in terms of KL divergence. Let $p^{(\text{best})}$ be the teacher checkpoint with the best performance. We reformulate Metric 1 as the KL divergence between $p^{(\text{best})}$ and $p^{(j)}$. This leads to the unified objective:

\[
j^* = \arg\min_{j \in N} \Big[ \mathop{\mathbb{E}}_{x \sim \mathcal{D}} \text{KL}\big( p^{(\text{best})}(x) \,\|\, p^{(j)}(x) \big) + \mathop{\mathbb{E}}_{x \sim \mathcal{D}} \text{KL}\big( f_s^{(i-1)}(x) \,\|\, p^{(j)}(x) \big) \Big].
\]

Here, the first term encourages selection of high‑performance teachers (close to the best teacher), and the second term enforces proximity to the current untrained student state, i.e., $f_s^{(i-1)}$.

\section{Adaptive Weight}
In domain-specific tasks, whether a student can surpass its teacher depends not only on optimization but also on task characteristics—specifically, the relative dominance of SFS and TFS, which vary across tasks. To actively shape this balance, we introduce the AW mechanism. By estimating SFS/TFS distributions from domain fine-tuned models, AW dynamically assigns higher distillation weight to TFS samples and lower weight to SFS ones. This focuses student learning on areas needing improvement while preserving its existing strengths.
\subsection{Metric for Adaptive Weight}

Let $p_T$ and $q_s$ denote the teacher and student model distributions after supervised fine-tuning on the domain dataset $\mathcal{D}$. For each sequence $\mathbf{y} \sim \mathcal{D}$, we define the autoregressive cross-entropy losses:

\[
\mathcal{L}_S(\mathbf{y}) = -\sum_{s=1}^{S} \log q_s(y_s \mid \mathbf{y}_{<s}), \quad 
\mathcal{L}_T(\mathbf{y}) = -\sum_{s=1}^{S} \log p_T(y_s \mid \mathbf{y}_{<s}),
\]

where $q_s(y_s \mid \mathbf{y}_{<s})$ and $p_T(y_s \mid \mathbf{y}_{<s})$ are the student and teacher probabilities for token $y_s$ given the preceding context $\mathbf{y}_{<s}$.

\begin{figure}[t]
\includegraphics[width=\textwidth]{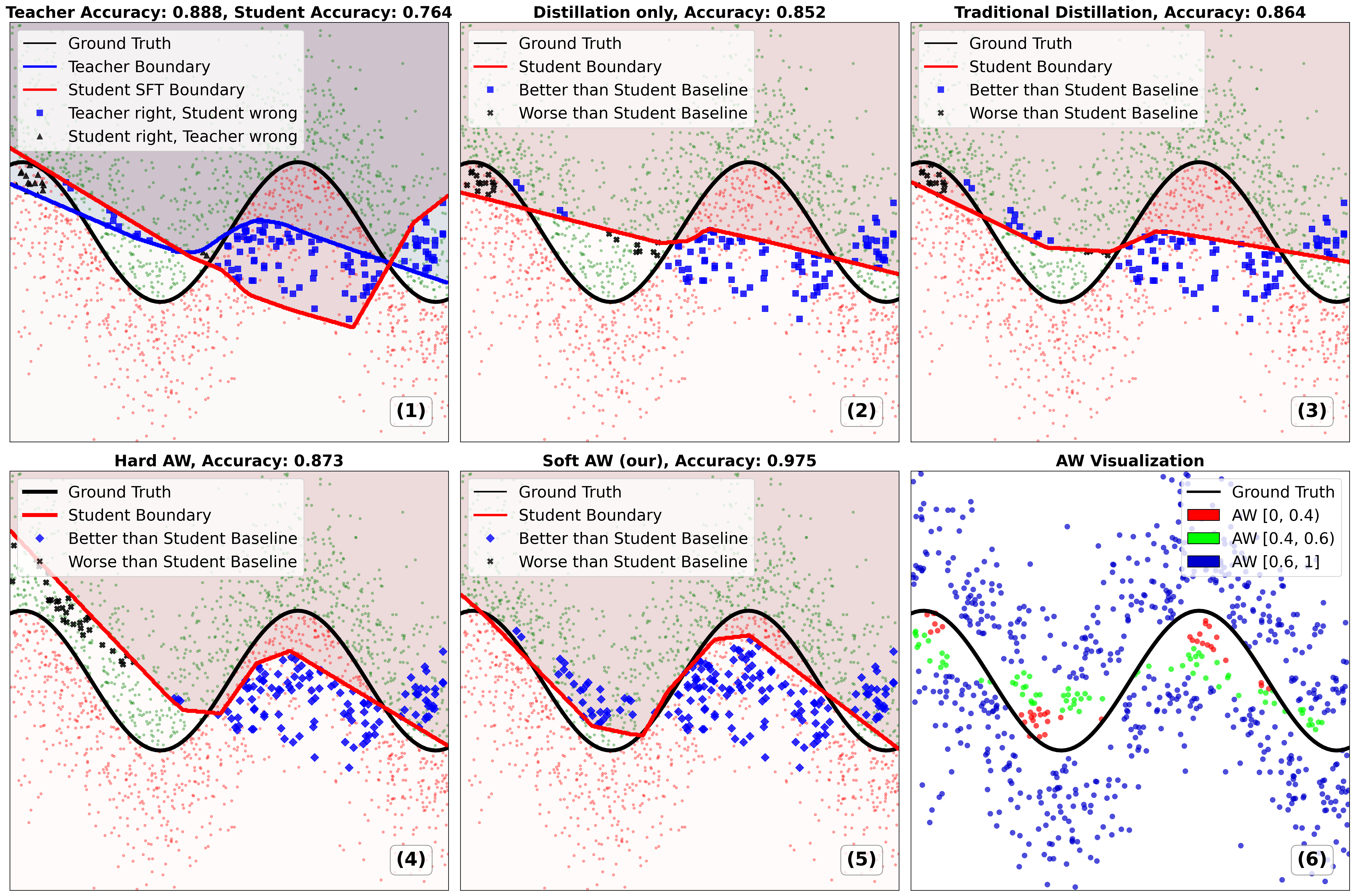}
\caption{\textbf{Toy Experiment for AW Visualization}. 
Data generated from sine function boundary $y=\sin(0.5x)+\mathcal{N}(0,1)$. 
Teacher model: 2-layer MLP (128 hidden units, $\sim$16K params). 
Student models: 2-layer MLP (8 hidden units, $\sim$50 params). 
(1) SFT Baselines. 
Distillation variants: (2) \textit{Distill Only}; (3) \textit{TD}: $0.5\mathcal{L}_{CE} + 0.5\mathcal{L}_{KL}$; 
(4) \textit{Hard AW}: teacher-better: only $\mathcal{L}_{KL}$, student-better: only $\mathcal{L}_{CE}$, others: TD; (5) \textit{Soft AW}: weighting based on loss ratio $\sigma(\log(\mathcal{L}_S/\mathcal{L}_T))$. (6) AW visualization.} \label{figure3}
\end{figure}

The sequence-wise adaptive weight $AW(\mathbf{y})$ is defined as:

\[
AW(\mathbf{y}) = \sigma \left( \log \frac{\mathcal{L}_S(\mathbf{y})}{\mathcal{L}_T(\mathbf{y})} \right),
\]

where $\sigma(\cdot)$ is the sigmoid function. The proposed method enforces a [0,1] weight range and monotonic correlation with the TFS/SFS distribution via log-scaled loss ratios and a sigmoid function. The total distillation objective becomes:

\[
\mathcal{L}_{\text{KD}}(\mathbf{y}) = AW(\mathbf{y}) \cdot \mathcal{L}_{\text{distill}}(p_T, q_\theta)  + \bigl(1 - AW(\mathbf{y})\bigr) \cdot \mathcal{L}_{\text{CE}}(\mathbf{y}, q_\theta) ,
\]

where $\mathcal{L}_{\text{CE}}(\mathbf{y}, q_\theta) = -\sum_{s=1}^{S} \log q_\theta(y_s \mid \mathbf{y}_{<s})$ is the standard cross-entropy loss with the ground-truth sequence, and $\mathcal{L}_{\text{distill}}(p_T, q_\theta)$ is the distillation loss (e.g., KL divergence) between $p_T$ and $q_\theta$. 
Figure~\ref{figure3} presents a toy experiment for visualizing AW. AW enhances the distillation loss for TFS Samples. Simultaneously, it reduces the distillation loss for SFS Samples, better preserving the student's strengths. This approach achieves superior performance compared to traditional distillation methods and significantly surpasses the teacher's performance (from 88.8 to 97.5).

\subsection{Apply AW to SCD}
Within the SCD framework, at each step $i$, the student model $f_s^{(i)}$ aligns with a selected teacher checkpoint $p^{(j)}$ from $\mathcal{C}_T = \{ p^{(1)}, \dots, p^{(N)} \}$. Since the teacher checkpoint $p^{(j)}$ may differ from the final fine-tuned teacher $p_T$, the original AW metric should also be adapted accordingly.

Specifically, instead of comparing the student's loss to that of the final teacher $p_T$, we compute the adaptive weight based on the student's SFT model $f_s^{(\text{SFT})}$ (which remains fixed throughout distillation) and the currently selected checkpoint $p^{(j)}$. For each input sequence $\mathbf{y} \sim \mathcal{D}$, we define:

\[
\mathcal{L}_S(\mathbf{y}) = -\sum_{s=1}^{S} \log f_s^{(\text{SFT})}(y_s \mid \mathbf{y}_{<s}), \quad 
\mathcal{L}_T^{(j)}(\mathbf{y}) = -\sum_{s=1}^{S} \log p^{(j)}(y_s \mid \mathbf{y}_{<s}),
\]

where $f_s^{(\text{SFT})}(y_s \mid \mathbf{y}_{<s})$ and $p^{(j)}(y_s \mid \mathbf{y}_{<s})$ denote the token‑level probabilities of the student SFT model and the $j$‑th teacher checkpoint, respectively.

The adaptive weight for step $i$ with teacher $p^{(j)}$ is then:

\[
AW^{(i,j)}(\mathbf{y}) = \sigma \left( \log \frac{\mathcal{L}_S(\mathbf{y})}{\mathcal{L}_T^{(j)}(\mathbf{y})} \right),
\]

Using this adaptive weighting, the final distillation loss for step $i$ with teacher checkpoint $p^{(j)}$ is formulated as:

\[
\mathcal{L}_{\text{KD}}^{(i,j)}(\mathbf{y}) = AW^{(i,j)}(\mathbf{y}) \cdot \mathcal{L}_{\text{distill}}\big(p^{(j)}, f_s^{(i)}\big) + \bigl(1 - AW^{(i,j)}(\mathbf{y})\bigr) \cdot \mathcal{L}_{\text{CE}}\big(\mathbf{y}, f_s^{(i)}\big),
\]

\section{Experiment}
\subsection{Experimental Setup}
\textbf{Datasets and Evaluation.} We evaluate on two domain-specific benchmarks in distinct languages: \textbf{PubmedQA} (English) and \textbf{JMED-LLM}\footnote{\url{https://github.com/sociocom/JMED-LLM}} (Japanese). For PubmedQA, we adopt its official 500/500 train/test split. For the JMED-LLM suite, we utilize five tasks: JMMLU (medical QA), CRADE (adverse event classification), RRTNM (cancer staging), SMDIS (disease classification), and NRNER (medical NER). Each JMED task follows an 80\%/20\% train/test split. To ensure pipeline compatibility, all tasks are formatted into a unified instruction-following structure following the ShareGPT template, implemented within the DistillKit framework\footnote{\url{https://github.com/arcee-ai/DistillKit}}. Evaluation is conducted on held-out test sets: PubmedQA uses Accuracy while JMED tasks are evaluated using the official toolkit with task-specific metrics, i.e., Exact F1/Partial F1 for \textit{NRNER} and Accuracy for others. Across all tasks, the student model checkpoint with the best performance is selected for comparison.

\textbf{Compared Baselines:} 
\textbf{TD~\cite{hinton2015distilling}:} The standard approach minimizing forward KL divergence.
\textbf{TAID~\cite{shing2025taid}:} Constructs a curriculum where the student matches a logit-interpolated target.
\textbf{CD~\cite{panigrahi2024progressive}:} Sequentially trains the student on a fixed schedule of intermediate checkpoints saved during teacher SFT.

\textbf{Training Configuration}. We employ \textbf{Llama-3.1-8B-Instruct} as teacher and \textbf{Llama-3.2-3B-Instruct} as student. Training consists of two stages:

\textbf{Domain-specific SFT:} Both teacher and student are fine-tuned on the domain-specific tasks. Key hyperparameters are tuned for the teacher model and applied identically to the student. $N$ evenly-spaced intermediate teacher checkpoints are saved during this stage.

\textbf{Task-specific Distillation:} For fair comparison, all methods share the same CD infrastructure and identical hyperparameters within each task: optimizer reset every $T$ steps, total $N \times T$ steps. Methods differ in teacher signal:
\begin{itemize}
\item \textbf{KD \& TAID}: Use only the best teacher checkpoint $p^{(best)}$.
\item \textbf{CD}: Follows fixed $(N,T)$-progressive schedule.
\end{itemize}

\textbf{Hyperparameter Settings.} We use the AdamW optimizer with $\beta_1=0.9$, $\beta_2=0.95$, and gradient clipping at 1.0. The learning rate schedule follows cosine decay with warmup (ratio=0.1). Max sequence length=512, batch size=8. Learning rate is tuned per task to ensure a strong TD baseline and $N$ is set by equally dividing the teacher model's training epochs across tasks . $T$ is fixed at 2 epochs for all tasks. We first establish a strong TD baseline via grid search, then apply the optimal configuration to all methods for controlled comparison. All experiments are conducted on NVIDIA A800 80GB GPUs.

\begin{table}[t]
\scriptsize
\setlength{\tabcolsep}{3pt} 
\caption{Comparison of various methods on multiple tasks. The highest score in each column is highlighted in bold.}
\centering
\begin{tabular}{@{}lccccccc@{}}
\toprule
Task Type & \multicolumn{2}{c}{QA} & \multicolumn{1}{c}{NER} & \multicolumn{3}{c}{Text Categorization} &  \\ 
\cmidrule(lr){2-3} \cmidrule(l){4-4} \cmidrule(lr){5-7}
Task Name & \makebox[1.1cm]{JMMLU} & \makebox[1.1cm]{PubmedQA} & \makebox[1.1cm]{NRNER} & \makebox[1.1cm]{CRADE } & \makebox[1.1cm]{RRTNM} & \makebox[1.1cm]{SMDIS} & \makebox[1.1cm]{Avg} \\
\midrule
Teacher zero-shot & 0.453 & 0.752 & 0.251/0.443 & 0.144  & 0.554 & 0.637 & 0.462 \\
Teacher SFT & 0.547 & 0.792 & 0.667/0.932 & 0.813  & 0.677 & 0.985 & 0.773 \\
Student zero-shot & 0.358 & 0.702 & 0.247/0.476 & 0.092  & 0.200 & 0.255 & 0.333 \\
Student SFT & 0.394 & 0.738 & 0.533/0.806 & 0.629  & 0.508 & 0.976 & 0.655 \\
TD~\cite{hinton2015distilling} & 0.453 & 0.750 & 0.676/0.894 & 0.804  & 0.523 & 0.986 & 0.727 \\
TAID~\cite{shing2025taid} & 0.504 & 0.762 & 0.659/0.866 & 0.804  & 0.523 & \textbf{\underline{0.988}} & 0.729 \\
CD~\cite{panigrahi2024progressive} & 0.482 & 0.754 & 0.684/0.883 & 0.801  & 0.585 & 0.986 & 0.739 \\
\rowcolor{gray!20}
SCD (OUR) & 0.474 & 0.756 & 0.686/0.909 & \textbf{\underline{0.819}}  & 0.538 & 0.986 & 0.742 \\
\rowcolor{gray!20}
SCD w/ AW (OUR) & \textbf{\underline{0.523}} & \textbf{\underline{0.766}} & \textbf{\underline{0.711/0.944}} & 0.807 & \textbf{\underline{0.600}} & 0.986 & \textbf{\underline{0.763}} \\
\bottomrule
\end{tabular}
\label{results}
\end{table}

\subsection{Overall results}

\begin{figure}[t]
    \centering
    \includegraphics[width=\textwidth]{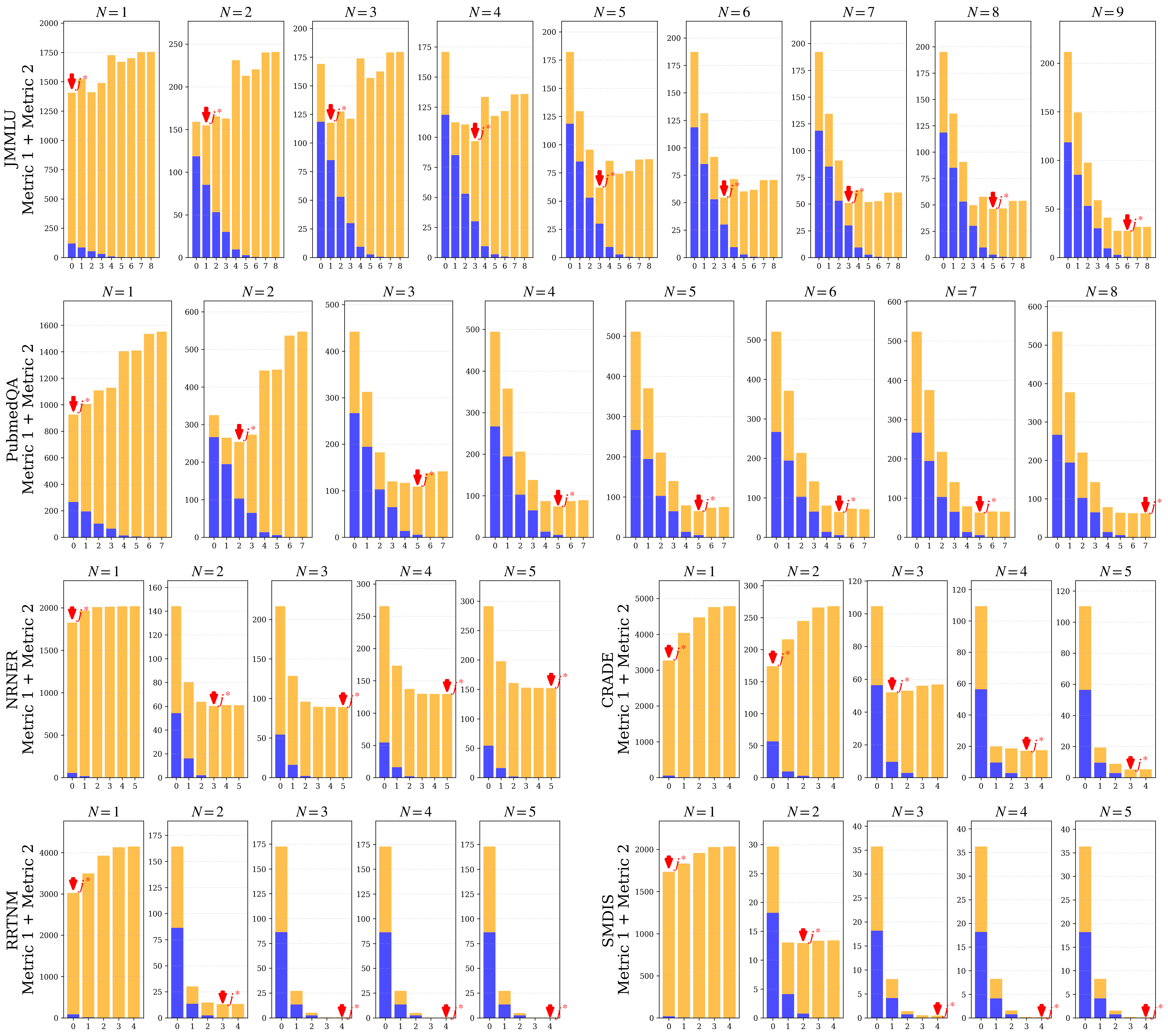}
    \caption{Schedule visualization based on the SCD w/ AW method. The blue and orange bars represent Metric 1 and Metric 2 (introduced in Section 4.2), respectively. The horizontal axis indicates the teacher checkpoint ID, while the vertical axis shows the sum of Metric 1 and Metric 2. Red arrows mark the optimal teacher selected.}
    \label{fig:scd_schedule}
\end{figure}

\begin{figure}[h]
\includegraphics[width=\textwidth]{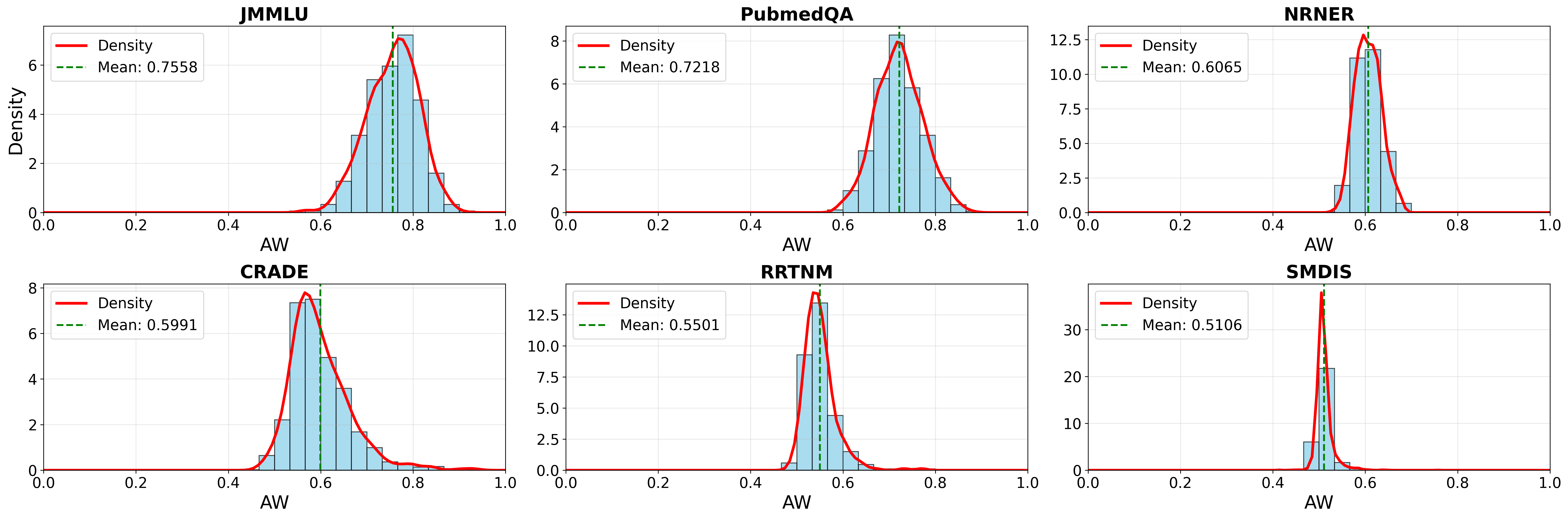}
\caption{\textbf{AW Visualization.} This demonstrates the distribution of AW on the training set, based on teacher SFT and student SFT.
} \label{figure4}
\end{figure}

As shown in Table~\ref{results}, our proposed methods consistently outperform existing distillation techniques. SCD achieves competitive performance (Avg: 0.742), while SCD w/AW demonstrates superior effectiveness (Avg: 0.763), excelling in key tasks such as JMMLU (+4.7\% vs CD) and NRNER Partial F1 (+6.9\% vs best baseline). Notably, SCD w/AW enables the student to surpass the teacher's performance in multiple tasks: it outperforms the teacher SFT model in both NRNER metrics (Exact F1: 0.711 vs 0.667, Partial F1: 0.944 vs 0.932) and matches or exceeds teacher performance in SMDIS (0.986 vs 0.985).

\subsection{Teacher Checkpoint Schedule Visualization}
Figure~\ref{fig:scd_schedule} visualizes the Teacher Checkpoint Schedule generated by our proposed \textbf{SCD w/ AW} method. The core algorithmic principle of our approach is to dynamically balance two complementary metrics: \textbf{Metric 1}, which prioritizes the teacher checkpoint with the highest task performance (i.e., the "best" teacher), and \textbf{Metric 2}, which favors the checkpoint that is most proximal or easiest for the current student to approximate (i.e., the "most teachable" teacher).

Our visualization and experimental results across all settings reveal a consistent and interpretable dynamic in the schedule. During the initial training phases, when the student model is relatively weak and distant from the teacher's performance manifold, \textbf{Metric 2} (approximation ease) dominates the selection. As training progresses and the student's capacity strengthens—thereby converging closer to the teacher's capability—the influence of \textbf{Metric 1} (peak teacher performance) progressively increases.

\subsection{AW Visualization on Training Dataset}

\begin{figure}[h]
\includegraphics[width=\textwidth]{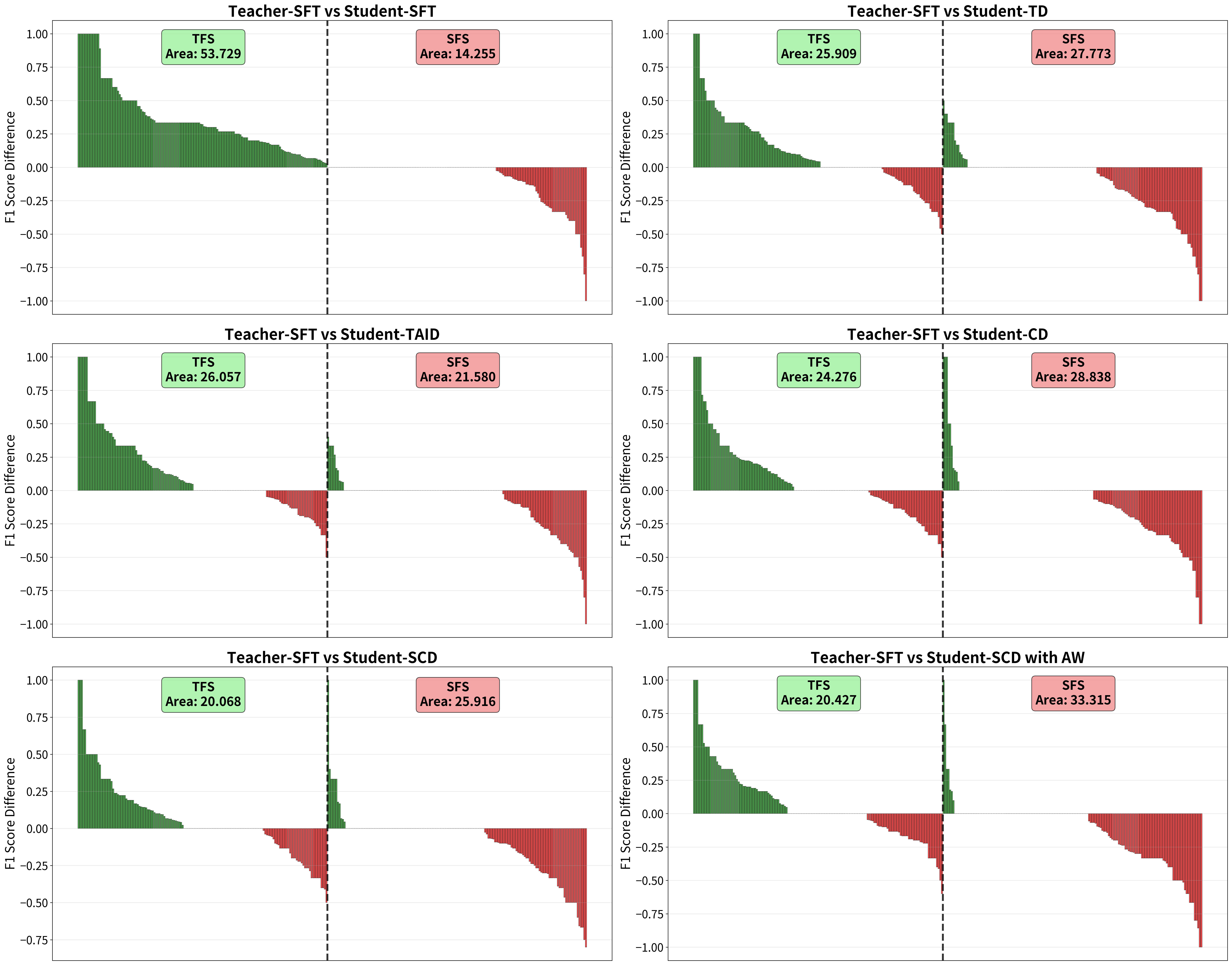}
\caption{\textbf{TFS \& SFS Visualization on NRNER.} In all subfigures, the vertical axis represents the sample-level F1 score difference (teacher – student), while the horizontal axis corresponds to the sample IDs. All plots are sorted in descending order along the vertical axis. SCD outperformed other methods in mitigating TFS, and the addition of AW successfully preserved SFS.
} \label{TFS}
\end{figure}
The AW mechanism we propose aims to elucidate the inherent advantage difference between the teacher and student models across different data samples. This discrepancy is largely task-specific and cannot be directly optimized as an objective. As showin in Figure~\ref{figure4}, on tasks such as JMMLU and PubmedQA, the teacher model possesses a substantial inherent advantage, making it difficult for any method to match, let alone exceed, the teacher’s performance. In contrast, on NRNER, CRADE, and SMDIS, our proposed method consistently outperforms the teacher. This improvement is primarily attributed to a higher proportion of samples in the AW distribution favoring the student in these tasks. 

\subsection{TFS \& SFS Visualization on Testing Dataset}

As illustrated in Figure~\ref{TFS}, we present visualizations of TFS and SFS for all baseline methods on the NRNER task. Among all compared approaches, our proposed SCD method achieves the lowest error rate on TFS, reducing the teacher model's lead from 53.729 to 20.068, thereby effectively alleviating the TFS deficit. By integrating the AW mechanism, our SCD w/ AW method further attains a significantly higher performance on SFS than all other baselines, reaching a maximum score of 33.315. Meanwhile, in the TFS domain, only a minor degradation is observed---from 20.068 to 20.427. Overall, our method enhances SFS performance while significantly reducing the TFS deficit, improving overall performance from 0.667 to 0.711.

\section{Conclusion}

In this paper, we explore a fundamental question: when and how can a student model match or surpass its teacher in domain-specific LLM distillation. Through theoretical analysis, we establish that this is achievable when the student's advantage on its favorable subdomain (SFS) outweighs its deficit on the teacher-favored subdomain (TFS). Guided by this insight, we introduce \textbf{SCD} to systematically reduce the TFS deficit by emulating the teacher's training trajectory, and \textbf{AW} to preserve the student's inherent strengths on SFS. Extensive experiments across diverse multilingual tasks show that our method surpasses existing distillation techniques and often allows the student model to match or even exceed the teacher’s performance. Our work provides both a theoretical foundation and a practical framework for developing efficient, high-performance domain-specific language models.

%
%
%
\bibliographystyle{splncs04}
%

\end{document}